\documentclass{article}
\usepackage[final]{neurips_2025}		

\usepackage{amsmath,amsfonts,bm}

\usepackage{blindtext}
\usepackage{subcaption}

\usepackage{xcolor}

\def\eqref#1{equation~\ref{#1}}

\def\1{\bm{1}}

\def\ve{{\bm{e}}}

\def\vr{{\bm{r}}}

\def\vv{{\bm{v}}}

\def\vx{{\bm{x}}}

\def\mA{{\bm{A}}}

\def\mC{{\bm{C}}}
\def\mD{{\bm{D}}}
\def\mE{{\bm{E}}}

\def\mL{{\bm{L}}}

\def\mX{{\bm{X}}}

\DeclareMathAlphabet{\mathsfit}{\encodingdefault}{\sfdefault}{m}{sl}
\SetMathAlphabet{\mathsfit}{bold}{\encodingdefault}{\sfdefault}{bx}{n}

\def\gE{{\mathcal{E}}}

\def\gG{{\mathcal{G}}}

\def\gR{{\mathcal{R}}}

\def\gX{{\mathcal{X}}}

\def\sR{{\mathbb{R}}}

\usepackage[utf8]{inputenc} 
\usepackage[T1]{fontenc}    
\usepackage{hyperref}       
\usepackage{url}            
\usepackage{booktabs}       
\usepackage{subcaption}
\usepackage{amsfonts}       
\usepackage{amssymb}
\usepackage{nicefrac}       
\usepackage{microtype}      
\usepackage{xcolor}         
\usepackage{makecell}

\usepackage{multirow}						
\usepackage{graphicx}						

\usepackage{comment}
\usepackage[ruled]{algorithm2e}
\usepackage{algorithmic,algorithm2e}

\makeatletter
\renewcommand{\@noticestring}{NeurIPS \@neuripsyear\ New Perspectives in Advancing Graph Machine Learning Workshop}
\makeatother

\title{LGDC: Latent Graph Diffusion \\ via Spectrum-Preserving Coarsening}

\author{%
  Nagham Osman\thanks{Core contributors}\\
  University College London\\
  London, UK \\
  \texttt{nagham.osman.21@ucl.ac.uk} \\
  \And
  Keyue Jiang$^*$ \\
  University College London \\
  London, UK \\
  \AND
  Davide Buffelli \\
  MediaTek Research \\
  London, UK \\
  \And
  Xiaowen Dong \\
  University of Oxford \\
  Oxford, UK \\
  \And
  Laura Toni \\
  University College London \\
  London, UK \\
}

\begin{document}

\maketitle

\begin{abstract}
Graph generation is a critical task across scientific domains. Existing methods fall broadly into two categories: autoregressive models, which iteratively expand graphs, and one-shot models, such as diffusion, which generate the full graph at once. In this work, we provide an analysis of these two paradigms and reveal a key trade-off: autoregressive models stand out in capturing fine-grained local structures, such as degree and clustering properties, whereas one-shot models excel at modeling global patterns, such as spectral distributions. Building on this, we propose LGDC (latent graph diffusion via spectrum-preserving coarsening), a hybrid framework that combines strengths of both approaches. LGDC employs a spectrum-preserving coarsening-decoarsening to bidirectionally map between graphs and a latent space, where diffusion efficiently generates latent graphs before expansion restores detail. This design captures both local and global properties with improved efficiency. Empirically, LGDC matches autoregressive models on locally structured datasets (Tree) and diffusion models on globally structured ones (Planar, Community-20), validating the benefits of hybrid generation.
\end{abstract}

\section{Introduction}
Graph generation underpins applications in drug discovery, molecular design, and social networks. Most approaches fall into two classes: (i) \emph{autoregressive models}, which construct graphs through iterative local expansion, and (ii) \emph{one-shot models}, which fix the graph size and generate the full structure in a single pass.

Early work on graph generation largely followed the autoregressive paradigm. GraphRNN~\cite{DBLP:conf/icml/YouYRHL18} pioneered this direction by modeling node interactions as a sequence of connection events. Advances in sequential modeling, particularly transformers, led to the emergence of a family of autoregressive graph generators, including GRAN~\citep{10.5555/3454287.3454670}, BiGG~\citep{dai2020scalable}, and GraphGen~\citep{goyal2020graphgen}. The most advanced system in this line of research is HSpectre~\citep{DBLP:conf/iclr/BergmeisterMPW24}, which frames generation as an iterative expansion process that grows a subgraph into a full graph; although architecturally distinct, each prediction step remains a local expansion, keeping it within the autoregressive family.

More recently, one-shot generative models~\citep{DBLP:conf/icml/JoLH22,DBLP:journals/corr/abs-2210-01549,DBLP:conf/iclr/VignacKSWCF23} have gained prominence. Key approaches include diffusion models~\citep{DBLP:conf/aistats/NiuSSZGE20, DBLP:conf/iclr/VignacKSWCF23,haefeli2022diffusion,xu2024discrete,siraudin2024comethcontinuoustimediscretestategraph} and flow-based models~\citep{DBLP:conf/nips/EijkelboomBNWM24,DBLP:journals/corr/abs-2410-04263, DBLP:journals/corr/abs-2411-05676, jiang2025bureswasserstein}, which generate entire graphs in a single shot, unlike autoregressive expansion. These models learn progressive transformations between a reference distribution and the data distribution by parameterizing the reverse denoising process with a neural network to enable sampling.

While both autoregressive and one-shot approaches have achieved considerable success in graph generation, a systematic understanding of their principles and trade-offs remains limited. This motivates the central question of this work:
\vspace{-0.6em}
\begin{center}
\emph{How do autoregressive and one-shot models compare, and can we combine their strengths?}
\end{center}
\vspace{-0.6em}

We hypothesize a trade-off between the two paradigms: autoregressive models, which generate graphs through sequential local expansions, excel at fine-grained local dependencies (e.g., parent–child links in Tree graphs) but often lose long-range coherence. Conversely, one-shot models learn holistic transformations of the adjacency structure, capturing global patterns, such as community structure or planarity, though often at the expense of local detail. This intuition guides our evaluation on datasets that stress different aspects: Tree graphs emphasize local attachment rules, while Planar and SBM graphs stress global organization. This intuition guides our evaluation on datasets that stress different aspects: Tree graphs emphasize local attachment rules, while Planar and SBM graphs stress global organization. This intuition is supported by Table~\ref{tab:local-global-datasets}, which summarizes training dataset characteristics across local and global metrics. Tree graphs are dominated by local dependencies, reflected in their stable local statistics and minimal global variation, whereas Planar and SBM graphs exhibit pronounced global organization in spectral and connectivity metrics, aligning with our hypothesis of local–global structural contrast. As shown in Table~\ref{tab:synthetic_graphs_vun_ratio}, autoregressive models perform best on Tree graphs, while one-shot models dominate on Planar and SBM graphs, confirming this trade-off.

\begin{table}[ht]
    \caption{Training Dataset Statistics across local and global metrics.}
    \label{tab:local-global-datasets}
    \centering
    \setlength{\tabcolsep}{2pt} 
    \renewcommand{\arraystretch}{1.02} 

    \resizebox{0.75\linewidth}{!}{ 
    \begin{tabular}{lccc@{\hspace{8pt}}lccc}
        \toprule
        \multicolumn{4}{c}{Local Metrics} & \multicolumn{4}{c}{Global Metrics} \\
        \cmidrule(r){1-4} \cmidrule(l){5-8}
        Metric & Tree & Planar & SBM &
        Metric & Tree & Planar & SBM \\
        \midrule
        Degree       & 0.0002 & 0.0000 & 0.0003 & Spectre     & 0.0091 & 0.0076 & 0.0060 \\
        Orbit        & 0.0000 & 0.0001 & 0.0310 & Components  & 0.0000 & 0.0000 & 0.0004 \\
        Motif        & 0.0000 & 0.0007 & 0.0346 & Edge Conn.  & 0.0000 & 0.0039 & 0.0205 \\
        Clustering   & 0.0000 & 0.0165 & 0.0331 & ASPL        & 0.0065 & 0.0000 & 0.0009 \\
        \addlinespace[0.5pt]
        & & & & Diameter    & 0.0241 & 0.0002 & 0.0001 \\
        \bottomrule
    \end{tabular}}
\end{table}

\begin{table*}[ht]
    \caption{Graph Generation Comparison. Autoregression models v.s. One-Shot generation. For a fair comparison, we disabled the training designs that are unrelated to the fundamental generation path mechanism, such as the target guidance DeFoG and the predictor-corrector in Cometh. Details regarding the evaluation metrics can be found in Section~\ref{sec: metrics}.}
    \label{tab:synthetic_graphs_vun_ratio}
    \vspace{-4pt}
    \centering
    \resizebox{1.0\linewidth}{!}{
    \begin{tabular}{lccccccc}
        \toprule
        & &  \multicolumn{2}{c}{  Planar } & \multicolumn{2}{c}{ Tree }& \multicolumn{2}{c}{ SBM } \\
        \cmidrule(lr){3-4} \cmidrule(lr){5-6} \cmidrule(lr){7-8}
        {Model} & {Class} & {V.U.N.\,\(\uparrow\)} & {A.Ratio\,\(\downarrow\)} & {V.U.N.\,\(\uparrow\)} & {A.Ratio\,\(\downarrow\)} & {V.U.N.\,\(\uparrow\)} & {A.Ratio\,\(\downarrow\)} \\
        \midrule
        {Train set} & {---} & {100} & {1.0} & {100} & {1.0} & {85.9} & {1.0} \\
        \midrule
        {BiGG~\citep{DBLP:conf/icml/DaiNLDS20}} & {Autoregressive} & 5.0 & 16.0 & 75.0 & 5.2 & 10.0& 11.9 \\
        {GraphGen~\citep{DBLP:conf/www/GoyalJR20}} & {Autoregressive} & 7.5 & 210.3 & 95.0 & 33.2 & 5.0 & 48.8 \\
        {HSpectre \citep{DBLP:conf/iclr/BergmeisterMPW24}} & {Autoregression} & 62.5 & 2.9 & 82.5 & 2.1 & 45.0 & 10.2 \\
        
        GruM~\citep{DBLP:conf/icml/JoKH24} & {One-shot} & {74.4\scriptsize{±5.15}} & 3.2\scriptsize{±0.4} & /& / & 73.5\scriptsize{±6.7}& {2.6\scriptsize{±0.6}} \\
        {Cometh~\citep{siraudin2024comethcontinuoustimediscretestategraph}} & {One-shot} & {75.5\scriptsize{± 7.37}} & {3.0\scriptsize{± 5.6}} & {69.5\scriptsize{± 3.6}} & 1.40\scriptsize{± 0.4} & 65.5\scriptsize{±4.5}& {4.7\scriptsize{±0.6}} \\
        {DeFoG~\citep{DBLP:journals/corr/abs-2410-04263}} & {One-shot} & {77.5\scriptsize{±8.37}} & {3.5\scriptsize{±1.7}} & 73.1\scriptsize{±11.4} & 1.50\scriptsize{±0.3} & 85.0\scriptsize{±7.1} & {3.7\scriptsize{±0.9}} \\
        \bottomrule
        \end{tabular}}
    \vspace{-4pt}
\end{table*}

\begin{figure}[t!]
\centering
  \begin{subfigure}[t]{0.48\textwidth}
    \centering
    \includegraphics[width=1\textwidth]{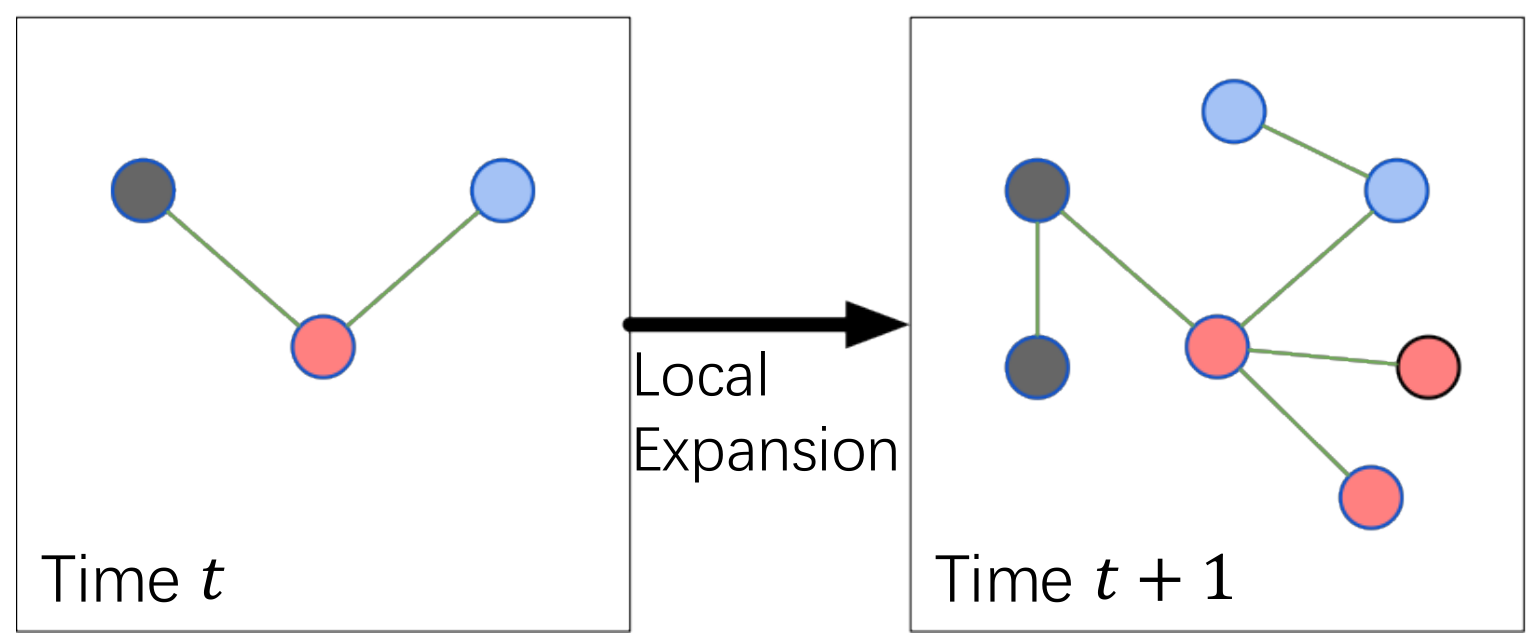}
    \caption{Graph generation with local expansion. The color represents the child nodes expanded from a same ancestral node.}
    \label{fig: gg_expand}
  \end{subfigure}
  \hfill
  \begin{subfigure}[t]{0.48\textwidth}
    \centering
    \includegraphics[width=1\textwidth]{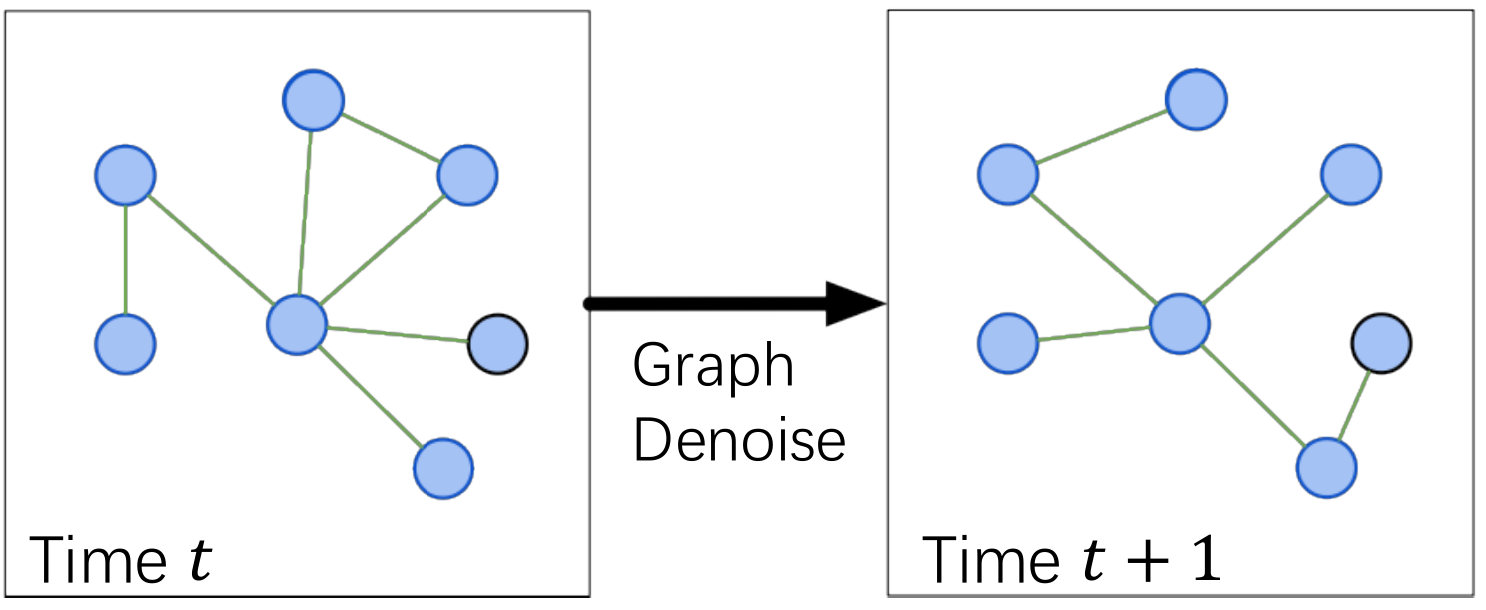}
    \caption{Graph Generation with iterative denoising. The number of nodes are fixed, where every individual node and edge have the possibility to change.}
    \label{fig: gg_diff}
  \end{subfigure}
  \caption{Two categories of graph generation methods.}
  \label{fig: gg_overview}
\end{figure}

Motivated by this complementarity, we propose a hybrid framework, \emph{latent graph diffusion via spectrum-preserving coarsening (LGDC)}\footnote{In this work, ‘latent space’ refers to the coarsened graph domain, not a Euclidean embedding.}. The intuition is as follows: one-shot diffusion captures global structure, but can be expensive on the original graph and may blur fine-grained connectivity. We instead run diffusion on a coarsened latent graph for efficient global modeling. To ensure global faithfulness, we adopt spectrum-preserving coarsening under restricted spectral similarity, which keeps the principal Laplacian eigenvalues and eigenspaces of the coarse and original graphs close~\citep{DBLP:conf/icml/LoukasV18}, allowing coarse eigenvectors to substitute for the originals, yielding a compact yet globally coherent surrogate. A single autoregressive-inspired expansion-and-refinement step during decoding then restores local connectivity. While Table~\ref{tab:synthetic_graphs_vun_ratio} does not directly evaluate our hybrid, it illustrates the motivating local--global trade-off: diffusion supplies global coherence, while autoregressive refinement sharpens local structure.

Concretely, LGDC follows a two-stage framework: a diffusion model first samples a small fixed-size latent graph $\gG_c$, which is then expanded back to a graph $\gG$ in the original space via a decoarsening model. Training leverages graph–coarsened graph pairs $(\gG, \gG_c)$ from spectrum-preserving coarsening, enabling joint optimization of the latent diffusion model $p_\theta(\gG_c)$ and expansion model $p_\phi(\gG \mid \gG_c)$.

Our contributions are summarized as follows:
\begin{itemize}
    \item We propose LGDC, a unified graph generation framework combining spectrum-preserving coarsening, latent diffusion, and a single expansion–refinement step for scalable generation.
    \item  LGDC captures both local (e.g., neighborhoods) and global (e.g., spectral) structures in one generative process. Empirically, it achieves results comparable to diffusion models on globally structured datasets (Planar, Community-20) and to autoregressive models on locally structured ones (Tree), demonstrating LGDC's strong balance between fine-grained accuracy and global coherence while maintaining high sample validity and diversity.
    \item LGDC achieves high efficiency and lower computational complexity compared to pure diffusion or autoregressive models (e.g., HSpectre). For graph size $n$, latent size $n_c$, and sampling steps $T$, its complexity is $O(n^2+Tn_c^2)$ versus $O(Tn^2)$ for one-shot and $O(Tn^2/3)$ for autoregressive generation assuming a same architecture for backbone models.
\end{itemize}

\section{Methodology}

\begin{figure}[t]
 \centering
 \includegraphics[width=0.7\textwidth]{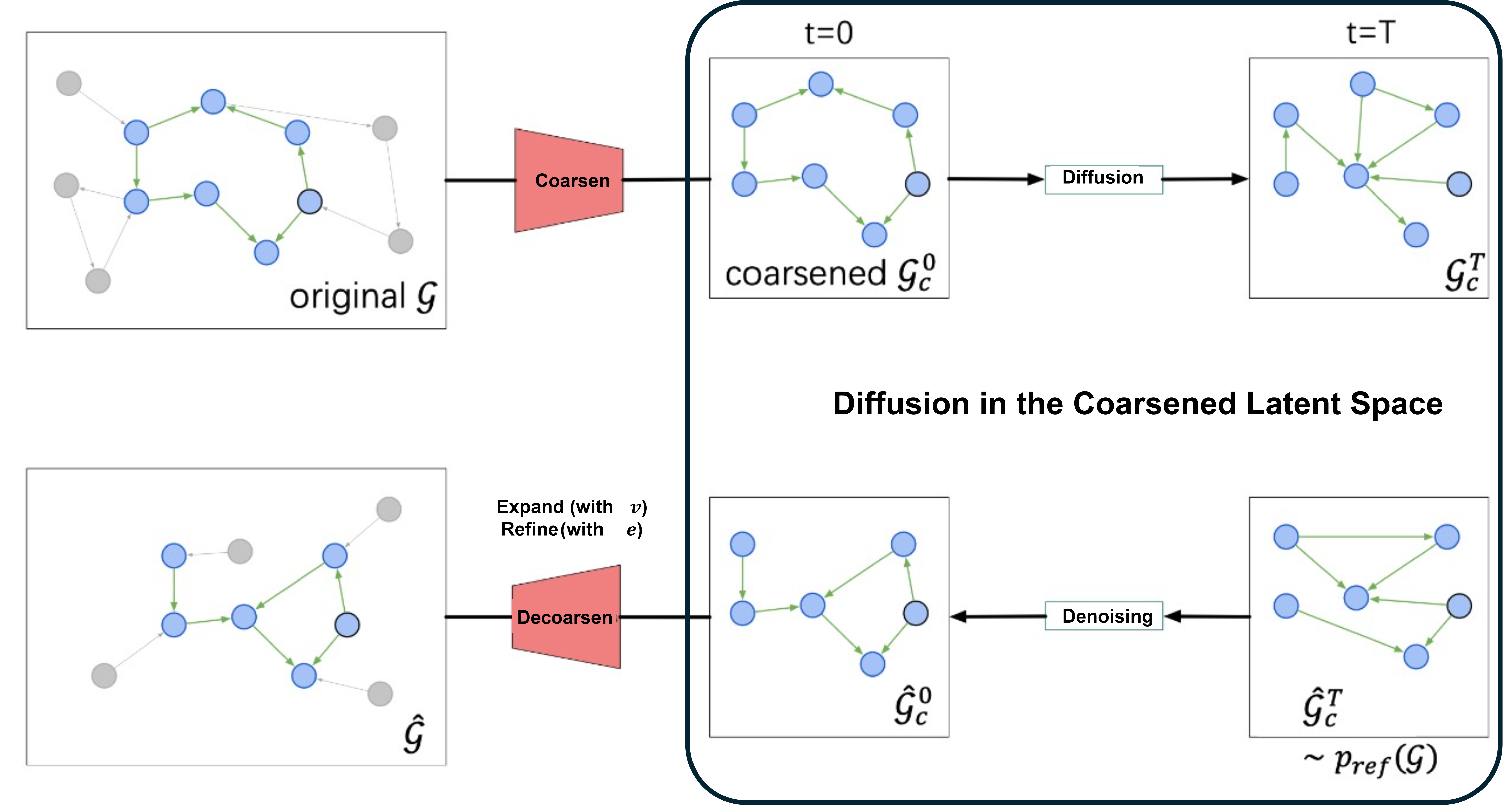}
 \caption{The architecture of LGDC framework (Latent Graph Diffusion via Coarsening)}
 \label{fig: coa_graph_diffusion}
\end{figure}

In this section, we introduce \emph{LGDC}, a hybrid graph generation model that first samples a graph in the latent space and expands it back to the original space. An overview is shown in Fig.~\ref{fig: coa_graph_diffusion} and Eq.~\ref{eq:pipeline}.
\begin{equation}
    \gG \xrightarrow{\text{coarsen }(\ref{eq:coarsen})} \gG_c
    \xrightarrow{\text{diffuse/denoise }(\ref{eq:forward})\text{--}(\ref{eq:reverse})} \hat{\gG}_c
    \xrightarrow{\text{one-shot expand/refine }(\ref{eq:expand-refine}),(\ref{eq:joint})} \hat{\gG}.
    \label{eq:pipeline}
\end{equation}

The subsequent sections present the core components of our approach: Subsection~\ref{subsec:problem} introduces the problem formulation, Subsection~\ref{subsec: coarsen} details the spectrum-preserving graph coarsening module, Subsection~\ref{subsec:diffusion} describes the latent-space diffusion process, Subsection~\ref{subsec:expand} outlines the parameterization of the expansion model, and Subsection~\ref{subsec: sampling} summarizes the sampling pipeline of the framework.

\subsection{Problem Formulation}
\label{subsec:problem}

We start by formalizing the problem. Let $\gG=(\mX,\mA)$ denote a graph with $n$ nodes, node features $\mX\in\mathbb{R}^{n\times d}$, and edge types encoded in $\mA\in\{0,1\}^{n\times n}$. 
The graph generation problem aims to approximate the data distribution $p(\gG)$ and sample novel graphs from it. Instead of modeling $p(\gG)$ directly in the original space, we introduce a coarsened latent graph representation $\gG_c=(\mX_c,\mA_c)$ of fixed size $n_c \ll n$. The generative process is parameterized by neural networks and expressed as:
\begin{equation}
    p(\gG) = \mathbb{E}_{p_\theta(\gG_c)} p_\phi(\gG \mid \gG_c),
\end{equation}
where $p_\theta(\gG_c)$ defines a distribution over latent graphs, and $p_\phi(\gG \mid \gG_c)$ maps them back to the original space.

\subsection{Spectrum-Preserving Graph Coarsening}
\label{subsec: coarsen}

Our approach leverages latent diffusion to model a graph’s global structure by operating on coarsened graphs that preserve key global properties of the original. We adopt spectrum-preserving coarsening under restricted spectral similarity~\citep{DBLP:conf/icml/LoukasV18}, which guarantees that the principal Laplacian eigenvalues and eigenspaces of the coarse and original graphs remain close. Since these spectral quantities encode fundamental topological characteristics such as connectivity, community structure, and planarity, the resulting coarse graph serves as a compact yet globally faithful surrogate~\citep{DBLP:conf/iclr/BergmeisterMPW24}.

Formally, given a graph $\gG$ with Laplacian $\mL=\mD-\mA$ and a projection matrix $\mC\in\{0,1\}^{n_c\times n}$ that maps each fine node to a coarse one, the coarsened graph $\gG_c = (\mX_c, \mA_c)$ is defined as
\begin{align}
    \mX_c &= \mC\,\mX, \quad \mA_c = \mD_c - \mL_c, \text{ with }
    \mL_c \;=\; \mC\,\mL\,\mC^\top. \label{eq:coarsen}
\end{align}
The coarse graph is spectrally $\epsilon$-approximate to the original if
\begin{equation}
(1-\epsilon)\,\mathrm{Tr}(\mX^{\top} \mL \mX) \;\leq\; \mathrm{Tr}(\mX_{c}^{\top} \mL_{c} \mX_{c}) \;\leq\; (1+\epsilon)\,\mathrm{Tr}(\mX^{\top} \mL \mX). 
\end{equation}
This restricted similarity criterion ensures the coarse graph preserves the spectral energy of the original, enabling it to mimic global structural patterns. To satisfy this condition, we adopt the REC algorithm~\citep{DBLP:conf/icml/LoukasV18}, a non-learning coarsening method in Algorithm~\ref{alg:rec}.

\subsection{Discrete Diffusion in the Latent Space }
\label{subsec:diffusion}

After coarsening the graphs, we lean on a discrete diffusion similar to~\cite{DBLP:conf/iclr/VignacKSWCF23} for parameterizing $p_\theta(\gG_c)$. Specifically, our goal is to build a diffusion model that transforms the samples from an easy-to-sample source distribution, \(\gG^T_c\sim p_T(\gG_c)\), to samples from the target distribution, \(\gG_c^0\sim p(\gG_c)\). This is done through defining an iterative sampling process, where
\begin{equation}
    p_\theta(\gG_c^0) = \int_{\gG_c^{[1:T-1]}} p(\gG_c^0\mid \gG_c^1)\cdot \ldots p(\gG_c^{t-1}\mid \gG_c^t)\cdot \ldots p(\gG_c^{T-1}\mid \gG_c^T)p(\gG_c^T) d\gG_c^{[1:T-1]}
\end{equation}
where we denote $d\gG_c^{[1:T-1]} \triangleq d\gG_c^{T-1}, \ldots  d\gG_c^{1}$. Thus, the model design translates to fitting a transition model, $p_\theta(\gG_c^{t-1}\mid \gG_c^t)$. To this end, we follow the permutation-equivariant graph transformer of~\citet{DBLP:conf/iclr/VignacKSWCF23} and decompose the transition model as,
\begin{align}
    &p_\theta(\gG_c^{t-1}\mid \gG_c^t) = p_\theta(\mX_c^{t-1} \mid \gG_c^{t}) \cdot  p_\theta(\mA_c^{t-1} \mid \gG_c^{t})\\
    &\text{with } p_\theta(\mX_c^{t-1} \mid \gG_c^{t}) 
    = \sum_{\mX^0_c} p\!\left(\mX_c^{t-1}\!\mid \mX_c^0, \mX_c^{t}\right)\,p_\theta(\mX^0_c \mid \gG_c^{t}), \\
    &\text{and }
    p_\theta(\mA_c^{(t-1)} \mid \gG_c^{(t)})= \sum_{\mA_c^0} p\!\left(\mA_c^{t-1}\!\mid \mA_c^0, \mA_c^{t}\right)\,p_\theta(\mA_c^0 \mid \gG_c^{t}).
    \label{eq:reverse}
\end{align}
where we parameterize the joint reconstruction probability by factorizing the node and edge predictions:
$p_\theta(\gG_c^0 \mid \gG_c^{t}) = p_\theta(\mX_c^0 \mid \gG_c^{t}) \cdot p_\theta(\mA_c^0 \mid \gG_c^{t})$. The model is trained to reconstruct the original graph $\gG_c^0$ from its noisy version $\gG_c^t$. We follow~\citet{DBLP:conf/iclr/VignacKSWCF23} and construct the noisy graph iteratively perturbing the clean graph through,
\begin{equation}
    p(\gG_c^{t} \mid \gG_c^{t-1}) \;=\; \big(X^{t-1} Q_X^{t},\, E^{t-1} Q_E^{t}\big), 
    \text{such that }
    p(\gG_c^{(t)} \mid \gG_c^0) \;=\; \big(X\,Q_X^{(1:t)},\, E\,Q_E^{(1:t)}\big),
    \label{eq:forward}
\end{equation}
where per–step transitions are defined by $Q_X^{(t)}\in[0,1]^{a\times a}$ and $Q_E^{(t)}\in[0,1]^{b\times b}$,which we refer to~\cite{DBLP:conf/iclr/VignacKSWCF23} for a detailed introduction.

\paragraph{Training objective.} The parameters $\theta$ are optimized to reconstruct the clean graph $\gG_c^0$ from its noisy counterpart $\gG_c^t$ at any timestep t. This is achieved by minimizing the combined cross-entropy loss for both node feature and adjacency matrix reconstruction.

\subsection{Expansion and Refinement}
\label{subsec:expand}

To reconstruct a full-resolution graph in a single decoding pass, we adopt the
\emph{expansion–refinement} formalism of \citet{DBLP:conf/iclr/BergmeisterMPW24}.
Let $\vv \in \mathbb{N}^{n_c}$ denote the \emph{node expansion vector}, where
$\vv_i$ specifies how many fine nodes are generated from each coarse node
$i$ in $\gG_c$, such that $\sum_{i=1}^{n_c} \vv_i = n$.
Let $\ve \in \{0,1\}^{|\tilde{\mathcal{E}}|}$ be an \emph{edge-selection mask}
over a set of candidate edges $\tilde{\mathcal{E}}$.

\paragraph{Expansion.}
The expansion operator $\tilde{\gG} = \mathrm{Expand}(\gG_c; \vv)$ replicates
each coarse node $i$ into $\vv_i$ fine nodes.
Candidate edges are then constructed in two groups:
(i) \textbf{intra-cluster edges}, connecting fine nodes that originate from the
same coarse node (modeling local substructure within a cluster);
and (ii) \textbf{inter-cluster edges}, connecting nodes that originate from
different coarse nodes $i$ and $j$ whenever $(i,j)$ is an edge in the coarse
graph $\gG_c$. In other words, inter-cluster connectivity is inferred from
the coarse-level adjacency, serving as a structural scaffold for generating
fine-level links between clusters. This process yields a dense candidate edge
set $\tilde{\mathcal{E}}$ used in the subsequent refinement step.

\paragraph{Refinement.}
The refinement step then applies the binary mask $\ve$ to prune edges and form
the final reconstructed graph $\hat{\gG}$:
\begin{equation}
    \gG_c \;\xrightarrow{\ \vv\ }\; \tilde{\gG} \;\xrightarrow{\ \ve\ }\; \hat{\gG}.
    \label{eq:expand-refine}
\end{equation}
Hence, $\ve_k = 1$ indicates that the $k$-th candidate edge in
$\tilde{\mathcal{E}}$ is retained in $\hat{\gG}$.

\paragraph{Parameterization and Training.}
Since $\gG$ is uniquely determined by $(\vv,\ve)$, the conditional model is
factorized as:
\begin{equation}
    p_\phi(\gG \mid \gG_c)
    = p_\phi(\vv, \ve \mid \gG_c)
    = p_\phi(\vv \mid \gG_c)\,
      p_\phi(\ve \mid \mathrm{Expand}(\gG_c; \vv), \gG_c).
    \label{eq:joint}
\end{equation}
The supervision pairs $(\vv^\star,\ve^\star)$ are obtained by inverting one
spectrum-preserving coarsening step.
Model parameters $\phi$ are optimized to maximize the reconstruction likelihood
(or equivalently the ELBO) following \citet{DBLP:conf/iclr/BergmeisterMPW24},
applied here to a single coarse-to-fine transition.

\subsection{Sampling Pipeline}
\label{subsec: sampling}
At test time, LGDC generates in two stages: (i) Latent diffusion starting from the prior at step $T$, we apply the DiGress-style reverse transitions (Eqs.~\ref{eq:forward}--\ref{eq:reverse}) to obtain a sampled latent graph $\hat{\gG}_c$, and (ii) One–shot decoding where we decode in a single coarse to fine pass by drawing $(\vv,\ve)\sim p_\phi(\cdot\,\mid \hat{\gG}_c)$ via Eq.~\ref{eq:joint} and applying Eq.~\ref{eq:expand-refine} to produce $\hat{\gG}$. Unlike \citet{DBLP:conf/iclr/BergmeisterMPW24}, which uses multiple expansion--refinement cycles (e.g., $1\!\to\!3\!\to\!8\!\to\!\cdots\!\to\!n$), LGDC performs a single coarse to fine inversion from $n_c$ to $n$ at training and testing that significantly reduce the computation complexity.

\paragraph{End-to-end Complexity.}
LGDC’s inference cost decomposes into (i) latent denoising on $\gG_c$, which is $O(T\,n_c^2)$, and (ii) one-shot expand/refine, which is $O(n^2)$ in the dense worst case. 
Thus the overall complexity is $
O\!\big(n^2 + T\,n_c^2\big)$. The sampling complexity calculation is in Appendix~\ref{apdx:complexity}.

\section{Experimental Results}

\subsection{Datasets and Evaluation Protocol}
\paragraph{Datasets.} We evaluate LGDC on three synthetic benchmarks that probe complementary structural regimes: (i) \emph{Tree} graphs~\citep{DBLP:conf/iclr/BergmeisterMPW24}, which are deterministic and acyclic, testing local structural fidelity; (ii) \emph{Planar} graphs~\citep{martinkus_spectre_2022}, which impose global planarity while still requiring consistent local layouts; and (iii) \emph{Community-20} graphs~\citep{DBLP:conf/iclr/VignacKSWCF23}, a small SBM dataset with connected clusters, challenging models to capture global community structure.

\paragraph{Setup.} 
We fix the coarsening ratio to approximately $n_c \approx n/5$ to balance fidelity and cost under a fixed compute budget. This ensures the latent graph remains large enough to preserve global structure while reducing the quadratic cost of diffusion in that space. Concretely, we set $M=16$ for Tree and Planar, and $M=4$ for Community-20. A broader sweep of coarsening ratios is left to future work.

\paragraph{Evaluation.} 
\label{sec: metrics}
We assess generated graphs on sample quality and structural fidelity. Table~\ref{tab:plain_graph_gen} reports validity, uniqueness, and novelty (V.U.N.), and the average ratio (A. Ratio), summarizing deviations of standard statistics from the reference distribution. To probe fidelity in detail, Table~\ref{tab:graphs_by_dataset_subtables} separates \emph{local} metrics (degree, clustering, orbit/motif), \emph{global} metrics (spectral distance, connectivity, diameter/ASPL), and \emph{cross-scale} wavelet statistics based on Laplacian eigen-decomposition, which combine both aspects. For compactness, we also report Local and Global Ratios, normalizing deviations against reference sets.

\subsection{Performance and Complexity Analysis}
\paragraph{Generation Performance.} 
The results in Table \ref{tab:plain_graph_gen} show that LGDC achieves the strongest performance on the locally-focused Tree dataset, reaching a V.U.N. score of 86.0 that surpasses even the specialized autoregressive HSpectre. On the globally oriented Planar and Community-20 graphs, LGDC also delivers competitive results: it improves V.U.N.\ on Planar to 82.5, outperforming both HSpectre and DeFoG, and attains consistently low discrepancy scores on Community-20 across statistics. These outcomes highlight LGDC’s strength as a hybrid model that unifies the complementary advantages of autoregressive and one-shot paradigms across both local and global generation tasks.

\paragraph{Generation Complexity. }
Standard one-shot graph generative models require $O(Tn^2)$ FLOPs, while autoregressive methods average $O(Tn^2/3)$. In contrast, LGDC operates much more efficiently, requiring only $
O\!\big(n^2 + T\,n_c^2\big)$ FLOPs, when $n_c \ll n$ (see Appendix~\ref{sec:complexity}). Since $T$ typically exceeds $n$ in graph generation tasks, the savings are substantial, and with small compression ratio $n_c / n$, LGDC achieves significant computational gains over both paradigms.

\begin{table*}[t]
    \caption{Graph Generation Results. We compare the most advanced autoregressive and one-shot methods and disable target guidance for fair comparison. The A.Ratio. result on Planar excludes Orbit ratios. The HSepctre results are taken from~\cite{DBLP:conf/iclr/BergmeisterMPW24} with one round of expansion.}
    \label{tab:plain_graph_gen}
    \centering
    \resizebox{1.0\linewidth}{!}{
    \begin{tabular}{lcccccccc}
        \toprule
        & &  \multicolumn{2}{c}{  Planar } & \multicolumn{2}{c}{ Tree }& \multicolumn{3}{c}{ Community-20 } \\
        \cmidrule(lr){3-4} \cmidrule(lr){5-6} \cmidrule(lr){7-9}
        {Model} & {Class} & {V.U.N.\,\(\uparrow\)} & {A.Ratio\,\(\downarrow\)} & {V.U.N.\,\(\uparrow\)} & {A.Ratio\,\(\downarrow\)} & {Deg.\,\ \(\downarrow\)} & {Clus.\,\ \(\downarrow\)} & {Orb.\,\ \(\downarrow\)}\\
        \midrule
        {HSpectre \citep{DBLP:conf/iclr/BergmeisterMPW24}} & {Autoregression} & 62.5 & \textbf{2.90} & 82.5 & 2.10 & / & / & / \\
        {DeFoG~\cite{DBLP:journals/corr/abs-2410-04263}} & {One-shot} & {77.5\scriptsize{±8.37}} & 4.07 & 73.1\scriptsize{±11.4} & \textbf{1.50} & 0.071 & 0.115 & 0.037 \\
        LGDC (Ours) & Hybrid & \textbf{82.5\scriptsize{±2.7}} & 3.06 & \textbf{86.0\scriptsize{±2.0}} &  1.70 & \textbf{0.037} & \textbf{0.027}& \textbf{0.007}\\
        \bottomrule
    \end{tabular}}
\end{table*}

\paragraph{Capturing Local and Global Patterns.}
Table~\ref{tab:graphs_by_dataset_subtables} provides a quantitative analysis of LGDC’s performance
before and after the expansion step. Each subtable corresponds to one dataset, while the two columns
(\textbf{Diffusion} vs.\ \textbf{Expansion}) represent the coarsened latent graphs ($\gG_c$) and
the reconstructed full graphs ($\hat{\gG}$), respectively. Metrics are grouped into
\textbf{Local}, \textbf{Global}, and \textbf{Local+Global} categories. Local metrics
(Degree, Clustering, Motif/Orbit) measure fine-grained connectivity; global metrics
(Spectral distance, Edge Connectivity, Components, Diameter, ASPL) reflect large-scale structure;
and joint metrics (Wavelet distance and V.U.N.) assess multi-scale consistency between both levels.

Numerically, the \emph{Local} metrics in all datasets show clear improvement after expansion, for
instance, degree and motif errors typically decrease by one order of magnitude
(e.g., from $0.0016$ to $0.0006$ or $0.0697$ to $0.0079$), confirming that the refinement step
successfully restores node-level structure and motif statistics.
The \emph{Global} metrics remain of the same scale before and after expansion, with small variations
in spectral or diameter scores (e.g., spectral distance changes from $0.0085$ to $0.0078$ or
$0.0126$ to $0.0090$), showing that the overall topological organization learned in the latent space
is largely preserved.
Meanwhile, the \emph{Wavelet} metric, which combines both local and global information, typically
decreases after expansion (e.g., $0.0078 \rightarrow 0.0041$ or $0.0089 \rightarrow 0.0063$),
indicating improved multi-scale coherence in the reconstructed graphs.
V.U.N.\ scores also remain high (above $80$ in all cases), verifying that sample validity and
diversity are retained.

Overall, these numerical patterns demonstrate that the latent diffusion step captures coherent global
structure, while the expansion–refinement step recovers fine-scale connectivity without sacrificing
global consistency. The complementary trends across all metric groups confirm LGDC’s ability to
produce graphs that are simultaneously globally organized and locally accurate within a single
decoding stage.


\begin{table*}[ht]
    \caption{Evaluation of LGDC across datasets. “Diffusion” refers to graphs in the coarsened latent space, and “Expansion” shows the same metrics after decoding to the original size.}
    \label{tab:graphs_by_dataset_subtables}
    \centering
    \setlength{\tabcolsep}{3pt}
    \renewcommand{\arraystretch}{1.15}

    \begin{subtable}[t]{0.32\linewidth}
        \caption{Tree}
        \centering
        \resizebox{\linewidth}{!}{
        \begin{tabular}{lcc}
            \toprule
            \textbf{Metric} & \textbf{Diffusion} & \textbf{Expansion} \\
            \midrule
            \multicolumn{3}{l}{\textbf{Local}} \\
            Degree      & 0.0005 {\scriptsize$\pm$ 0.0003} & 0.0002 {\scriptsize$\pm$ 0.0002} \\
            Orbit       & 0.0001 {\scriptsize$\pm$ 0.0000} & 0.0000 {\scriptsize$\pm$ 0.0000} \\
            Clustering  & 0.0000 {\scriptsize$\pm$ 0.0000} & 0.0000 {\scriptsize$\pm$ 0.0000} \\
            L. Ratio    & 4.22 {\scriptsize$\pm$ 3.11}      & 0.86 {\scriptsize$\pm$ 0.52} \\
            \addlinespace
            \multicolumn{3}{l}{\textbf{Global}} \\
            Spectre     & 0.0085 {\scriptsize$\pm$ 0.0012} & 0.0078 {\scriptsize$\pm$ 0.0009} \\
            Components  & 0.0003 {\scriptsize$\pm$ 0.0002} & 0.0088 {\scriptsize$\pm$ 0.0053} \\
            Diameter    & 0.0249 {\scriptsize$\pm$ 0.0120} & 0.0789 {\scriptsize$\pm$ 0.0202} \\
            G. Ratio    & 1.79 {\scriptsize$\pm$ 0.63}      & 2.21 {\scriptsize$\pm$ 0.31} \\
            \addlinespace
            \multicolumn{3}{l}{\textbf{Local + Global}} \\
            Wavelet     & 0.0078 {\scriptsize$\pm$ 0.0004} & 0.0041 {\scriptsize$\pm$ 0.0003} \\
            VUN         & 97.5 {\scriptsize$\pm$ 1.58}     & 86.0 {\scriptsize$\pm$ 2.0} \\
            \bottomrule
        \end{tabular}}
    \end{subtable}%
    \hfill
    \begin{subtable}[t]{0.32\linewidth}
        \caption{Planar}
        \centering
        \resizebox{\linewidth}{!}{
        \begin{tabular}{lcc}
            \toprule
            \textbf{Metric} & \textbf{Diffusion} & \textbf{Expansion} \\
            \midrule
            \multicolumn{3}{l}{\textbf{Local}} \\
            Degree      & 0.0016 {\scriptsize$\pm$ 0.0005} & 0.0006 {\scriptsize$\pm$ 0.0001} \\
            Clustering  & 0.0553 {\scriptsize$\pm$ 0.0020} & 0.0509 {\scriptsize$\pm$ 0.0133} \\
            Motif       & 0.0697 {\scriptsize$\pm$ 0.0131} & 0.0079 {\scriptsize$\pm$ 0.0030} \\
            L. Ratio    & 4.81 {\scriptsize$\pm$ 1.52}      & 19.88 {\scriptsize$\pm$ 3.51} \\
            \addlinespace
            \multicolumn{3}{l}{\textbf{Global}} \\
            Spectre     & 0.0126 {\scriptsize$\pm$ 0.0021} & 0.0090 {\scriptsize$\pm$ 0.0011} \\
            Edge Conn.  & 0.0162 {\scriptsize$\pm$ 0.0036}      & 0.0099 {\scriptsize$\pm$ 0.0030} \\
            Diameter    & 0.0047 {\scriptsize$\pm$ 0.0046} & 0.0106 {\scriptsize$\pm$ 0.0086} \\
            G. Ratio    & 9.03 {\scriptsize$\pm$ 6.81}      & 19.46 {\scriptsize$\pm$ 7.34} \\
            \addlinespace
            \multicolumn{3}{l}{\textbf{Local + Global}} \\
            Wavelet     & 0.0089 {\scriptsize$\pm$ 0.0008} & 0.0063 {\scriptsize$\pm$ 0.0005} \\
            VUN         & 100.0 {\scriptsize$\pm$ 0.0}     & 82.5 {\scriptsize$\pm$ 2.74} \\
            \bottomrule
        \end{tabular}}
    \end{subtable}%
    \hfill
    \begin{subtable}[t]{0.32\linewidth}
        \caption{Community-20}
        \centering
        \resizebox{\linewidth}{!}{
        \begin{tabular}{lcc}
            \toprule
            \textbf{Metric} & \textbf{Diffusion} & \textbf{Expansion} \\
            \midrule
            \multicolumn{3}{l}{\textbf{Local}} \\
            Degree      & 0.0060 {\scriptsize$\pm$ 0.0030} & 0.0369 {\scriptsize$\pm$ 0.0028} \\
            Clustering  & 0.0322 {\scriptsize$\pm$ 0.0193} & 0.0266 {\scriptsize$\pm$ 0.0023} \\
            Motif       & 0.0030 {\scriptsize$\pm$ 0.0012} & 0.0317 {\scriptsize$\pm$ 0.0059} \\
            L. Ratio    & 0.45 {\scriptsize$\pm$ 0.24}      & 1.92 {\scriptsize$\pm$ 0.69} \\
            \addlinespace
            \multicolumn{3}{l}{\textbf{Global}} \\
            Spectre     & 0.0457 {\scriptsize$\pm$ 0.0175} & 0.1431 {\scriptsize$\pm$ 0.0084} \\
            Components  & 0.0021 {\scriptsize$\pm$ 0.0016} & 0.0029 {\scriptsize$\pm$ 0.0017} \\
            ASPL        & 0.0022 {\scriptsize$\pm$ 0.0020}      & 0.0014 {\scriptsize$\pm$ 0.0012} \\
            G. Ratio    & 0.53 {\scriptsize$\pm$ 0.32}      & 1.18 {\scriptsize$\pm$ 0.14} \\
            \addlinespace
            \multicolumn{3}{l}{\textbf{Local + Global}} \\
            Wavelet     & 0.0410 {\scriptsize$\pm$ 0.0166} & 0.1175 {\scriptsize$\pm$ 0.0061} \\
            VUN         & ---      & --- \\
            \bottomrule
        \end{tabular}}
    \end{subtable}
\end{table*}

\section{Conclusion}
We presented \emph{LGDC}, a hybrid graph generator that combines spectrum-preserving coarsening with latent discrete diffusion and a single expansion--refinement step. By design, this approach unifies the complementary strengths of one-shot and autoregressive paradigms: diffusion in a compact latent space imposes global organization at low cost, while local-expansion based decoding restores fine local structure. Empirically, LGDC achieves strong V.U.N.\ and competitive discrepancy scores with an end-to-end sampling complexity of $O(n^2 + T n_c^2)$, offering practical efficiency. Nonetheless, several limitations remain. The one-shot expand–refine decoding can be error-prone: small inaccuracies in the edge mask $\ve$ may propagate when the candidate edge set $\tilde{\mathcal{E}}$ is large, occasionally disturbing coarse-level consistency. Although latent diffusion is efficient, the expansion step still scales as $O(n^2)$ in the dense case, making very large graphs challenging to handle. The model’s performance also depends on the chosen coarsening procedure and compression ratio $n_c/n$, as different projection matrices can yield variable cluster structures and reconstruction difficulty. Finally, the current refinement does not explicitly enforce global constraints such as planarity or connectivity, and our experiments focus on synthetic benchmarks rather than large, real-world graphs. Future work will explore iterative or uncertainty-aware refinement to mitigate decoding errors, sparse or structured expansion to improve scalability, and adaptive or learned spectrum-preserving coarsening for greater robustness. Extending LGDC to richer node and edge attributes and larger heterogeneous graphs offers a promising path toward scalable, structure-aware generative modeling across domains.

\bibliographystyle{unsrtnat}
\bibliography{ref}

\newpage
\appendix
\section{Complexity Comparison}
\label{apdx:complexity}
\subsection{Scalability in Graph Generation and the Computational Complexity Issue}
Even though both autoregressive and one-shot models have achieve satisfactory results in graph generation, the scalability has long been a problem due to the heavy computation budget. In this section, we will illustrate the computational complexity of both families of graph generation methods.

The time complexity, consisting of \textit{Training Complexity} and \textit{Inference Complexity}, is related to the number of the floating point operations (FLOPs) (e.g. plus/multiplication) in the forward-propagation and back-propagation (where the gradient is calculated). The inference complexity consists of making predictions through the model in each step. The space complexity mainly concerns the memory usage of a model (RAM or GPU memory). The complexity of time and space is highly intertwined and is measured through the model architecture. As a brief solution, calculating each step of propagation with a specific model architecture is good enough for measuring the complexity.

\paragraph{The computational complexity of one-shot diffusion/flow models.} The cost mainly comes from the fact that one-shot graph generation models uses a dense representation of graphs for computation, which results in a quadratic complexity w.r.t \#nodes (n) and \#dimension (d). The complexity of one-shot graph generation models, including Digress~\cite{DBLP:conf/iclr/VignacKSWCF23}, DeFog\cite{DBLP:journals/corr/abs-2410-04263} and BWFlow\cite{jiang2025bureswasserstein}, are generally similar as the recent development of this family of models mainly focusing on accelerating the sampling and reducing the training cost. The sampling steps $T$ is a parameter that we cannot easily quantify. Though we assume it to be fixed here for simplicity but we also wish to emphasize its importance in graph generation.  

\begin{figure}[ht]\centering
\includegraphics[scale=0.4]{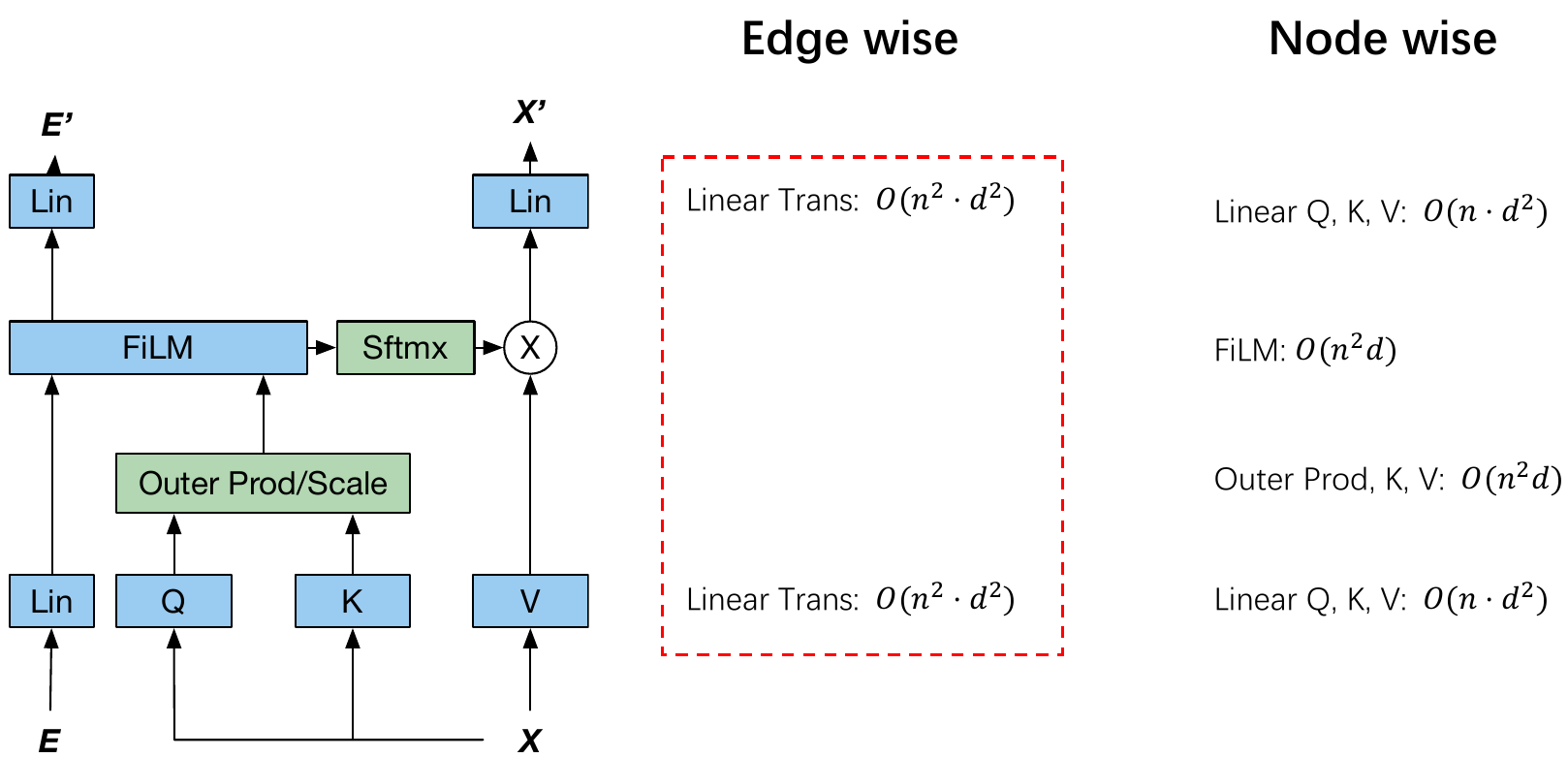}
\caption{The time complexity of One-shot graph generation. Given the number of nodes $n$, dimension size $d$, number of layers $L$, and number of samples $S$, the overall model complexity for training is given as $O(S\cdot L\cdot (n^2\cdot d^2+n^2\cdot d+n\cdot d^2))$. The inference complexity $O(T\cdot L\cdot (n^2\cdot d^2+n^2\cdot d+n\cdot d^2))$ for a single sample. The leading term is $n^2\cdot d^2$. which comes from the edge-wise prediction. }
\label{fig: model_time_complex}
\end{figure}

We consider the graph transformer~\cite{DBLP:conf/iclr/VignacKSWCF23} as the backbone model, which has a model architecture that is shown in~\ref{fig: model_time_complex} (simplified by removing all the conditional generation component). 
DiGress handles graphs with categorical node and edge attributes, represented by the spaces $\gX$ and $\gE$, respectively, with cardinalities $a$ and $b$. 
We use $x_i$ to denote the attribute of node $i$ and $\vx_i \in \gR^a$ to denote its one-hot encoding. These encodings are organized in a matrix $\mX \in \gR^{n \times a}$ where $n$ is the number of nodes. A tensor $\mE \in \sR^{n \times n \times b}$ groups the one-hot encoding $\vr_{i j}$ of each edge, treating the absence of edge as a particular edge type.  Each step of diffusion can be thought to be a joint node regression and link prediction process, where one would reverse the noising process of $q(\mE_t, \mX_t \mid \mE_{t-1}, \mX_{t-1})$, as $p_\theta(\mE_{t-1}, \mX_{t-1} \mid \mE_{t}, \mX_{t})$. The forward path progressively corrupts the data points and formalizes a sequence, namely $(\mX_0, \ldots, \mX_T)$ and $(\mE_0, \ldots, \mE_T)$.

In the reverse path, a backbone model parameterized by $\theta$ is used to learn the process, where we denote as $p_\theta: \gX \times \gE \rightarrow \gX \times \gE$. As an example, the continuous graph generation models require sample $n^2$, which yields a complexity of $O(n^2T)$. The discrete graph generation algorithm asks for a transition matrix that is capable of modeling a transition space of $\gR^{n^2} \times \gR^{n^2}$. Omitting the embedding size, the complexity is $O(n^2)$.

\paragraph{The computational complexity of auto-regressive models.} 

Assuming the backbone model to be the same graph generation, a single step of autoregressive computation follows a similar pipeline. However, the difference comes from the fact that each expansion steps the graph transformer operates on graphs with different size. We consider T step expansion and each step expands $(n-1)/T$ nodes. Omitting the hidden dimension and number of layers, we now aim to calculate 
\[S=\;1^2 + \Bigl(1+\frac{n-1}{T}\Bigr)^2 + \Bigl(1+2
\cdot \frac{n-1}{T}\Bigr)^2 + \cdots + n^2.
\]
Let \(d=\frac{n-1}{T}\)., we simplify to\(S=\sum_{t=0}^{T} \bigl(1+t d\bigr)^2.\). Expanding, we get
\((1+t d)^2 = 1 + 2t d + t^2 d^2,\)
so
\[
\begin{aligned}
S&=\sum_{t=0}^{T}\left(1 + 2t d + t^2 d^2\right) = \sum_{t=0}^{T}1 \;+\; 2d\sum_{t=0}^{T} t\;+\; d^2\sum_{t=0}^{T} j^2\\
&= \frac{T+1}{6T}\Bigl(6Tn + (2T+1)(n-1)^2\Bigr).\,
\end{aligned}
\]

Thus, the complexity is $O(
\frac{Tn^2}{3})$.

\subsection{The computational complexity of latent graph generation}
\label{sec:complexity}
 In the image generation task, a solution to generation is latent diffusion~\cite{DBLP:conf/cvpr/RombachBLEO22, DBLP:conf/nips/VahdatKK21}, where one first compares a high-resolution image into a low-rank latent space, and then diffuses over the latent space to achieve efficient sampling. 

 For this scenario, lets consider first compressing the graph into a latent space, with size $n_c$, we then operates on the latent space for $T/2$ step of generation, which yields $O(T/2\cdot L\cdot (n_c^2d^2))$ inference complexity. We then expand with one-step graph generation, which again has $O( T/2\cdot L\cdot (n_c^2d^2))$ FLOPs as the model only need to output the node expansion operator. Then, the edge predictor, which is a sparse graph neural network that takes in the expanded graph, which has worst $O( L\cdot (n^2d^2))$ complexity. Thus, omitting the embedding size and layer complexity, the total computation complexity is $O(n^2+Tn_c^2)$. Comparing to the full one-shot graph genertaion model ($O(Tn^2)$) and autoregressive model($O(Tn^2/3)$), the latent graph generation model has a significant gain in the computational efficiency as long as $\frac{n_c}{n}$ is small sufficiently.

\section{Related Work}
Hierarchical structures for graph generation have been explored in several prior works~\cite{DBLP:conf/aistats/GuoZL23, DBLP:journals/mlst/HyK23, DBLP:conf/iclr/BergmeisterMPW24}. \citet{DBLP:journals/mlst/HyK23} proposed a hierarchical VAE that progressively reduces graph size from $N_t$ to $N_{t+1}$ through soft cluster assignments, similar to DiffPool~\cite{DBLP:conf/nips/YingY0RHL18}. While effective for capturing multiscale structure, this approach assumes a fixed number of nodes per layer, is computationally expensive, and lacks theoretical guarantees on information preservation. On the other hand, \citet{DBLP:conf/iclr/BergmeisterMPW24} interpret the forward process of diffusion as graph coarsening and the reverse process as graph expansion. Their method iteratively adds and removes nodes and edges, bridging between a singleton graph and the full target distribution.  

To the best of our knowledge, only two works explicitly consider latent graph diffusion~\cite{10508504, DBLP:journals/corr/abs-2402-02518}. These frameworks apply diffusion in a latent space but do not compress graphs by reducing the number of nodes (from $n$ to $n_c$ with $n \gg n_c$). Instead, they embed graphs into a Euclidean vector space, which limits scalability and prevents fully leveraging the efficiency benefits of latent diffusion. In contrast, our model combines spectrum-preserving graph coarsening with latent diffusion in the non-Euclidean space of graphs and a single expansion–refinement stage. This design enables efficient sampling in a substantially smaller space while maintaining a balance between global structural organization and high-fidelity local reconstruction.

\section{Randomized edge contraction (REC) algorithm}

We introduce the REC algorithm in~\ref{alg:rec}.

\begin{algorithm}[h!]
\caption{Randomized Edge Contraction (REC)}
\label{alg:rec}
\SetAlgoLined
\KwIn{Graph $G=(\mX, \mA)$, iteration limit $T$.}
\KwOut{Coarsened graph $G_{\mathrm{c}}=\left(\mX_{c}, \mA_{c}\right)$.}
\BlankLine
$\mathcal{C} \leftarrow \mathcal{E}$, where $\gE$ is the edge set where $\mA_{ij}\neq0$\;
$G_{\mathrm{c}} \leftarrow G$\;
$\Phi \leftarrow \sum_{e_{i j} \in \mathcal{E}} \mA_{i j}$\;
$t \leftarrow 0$\;
$p_{\text{null}} \leftarrow 0$\;
\BlankLine
\While{$|\mathcal{C}|>0$ \text{ and } $t<T$}{
    $t \leftarrow t+1$\;
    Sample an outcome from $\mathcal{C} \cup \{\text{null}\}$ with probabilities $p_{ij}=\mA_{i j} / \Phi$ and $p_{\text{null}}$\;
    \If{an edge $e_{ij} \in \mathcal{C}$ was sampled}{
        $G_{\mathrm{c}} \leftarrow \operatorname{contract}(G_{\mathrm{c}}, e_{i j})$ \tcp*{as in Eq. (2)}
        Let $\mathcal{N}_{i j}$ be the set of edges incident to nodes $i$ or $j$\;
        $\mathcal{C} \leftarrow \mathcal{C} \setminus \mathcal{N}_{i j}$ \tcp*{Remove neighbors}
        $p_{\text{null}} \leftarrow p_{\text{null}} + \sum_{e_{pq} \in \mathcal{N}_{ij}} p_{pq}$\;
    }
}
\KwRet{$G_{\mathrm{c}}$}\;
\end{algorithm}

\section{Visualizations}
 \begin{figure*}[ht]
    \centering
    \begin{subfigure}[t]{0.28\linewidth}
        \centering
        \includegraphics[width=\linewidth]{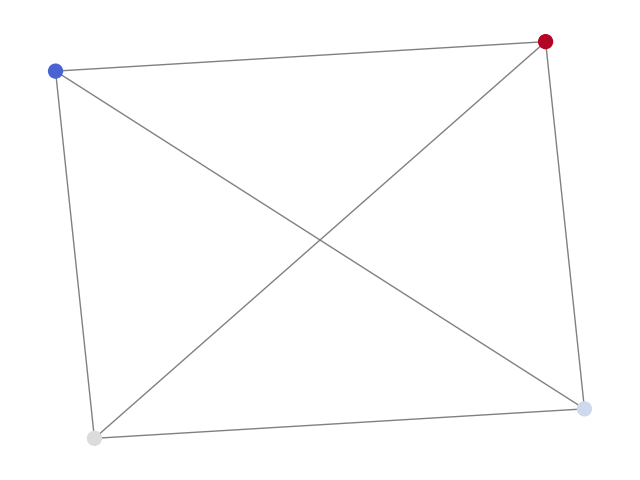}
        \caption{Comm20 (Noise)}
    \end{subfigure}\hfill
    \begin{subfigure}[t]{0.28\linewidth}
        \centering
        \includegraphics[width=\linewidth]{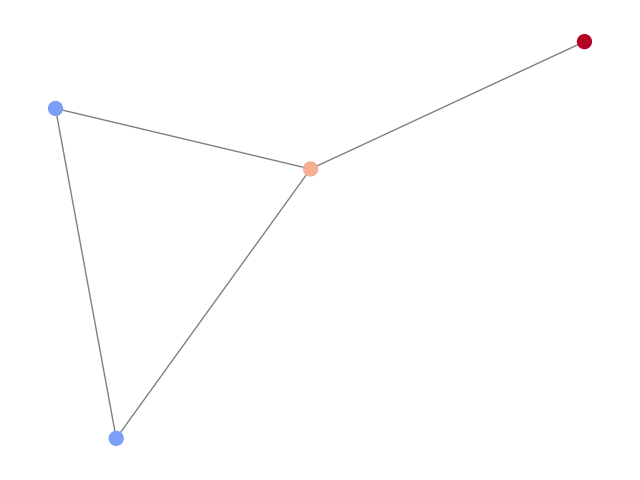}
        \caption{Comm20 (Coarsened)}
    \end{subfigure}\hfill
    \begin{subfigure}[t]{0.28\linewidth}
        \centering
        \includegraphics[width=\linewidth]{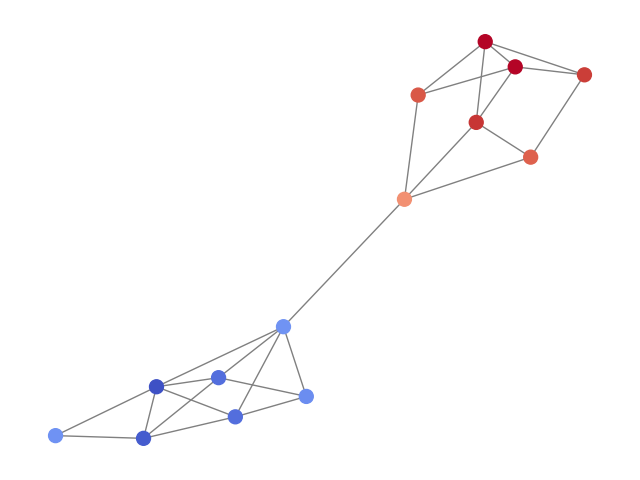}
        \caption{Comm20 (Expanded)}
    \end{subfigure}

    \vspace{0.4cm}
    \begin{subfigure}[t]{0.28\linewidth}
        \centering
        \includegraphics[width=\linewidth]{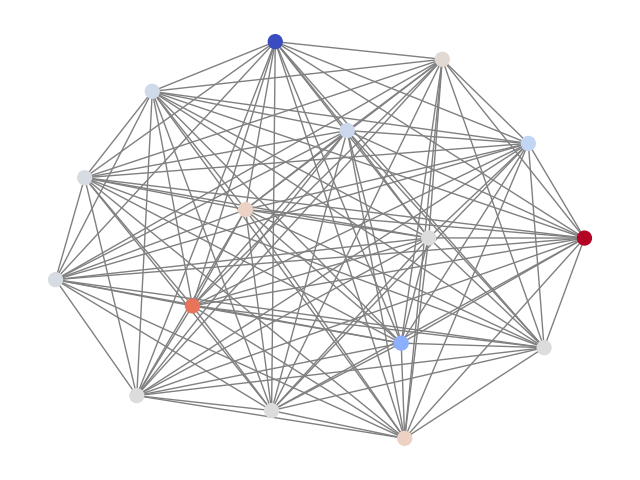}
        \caption{Tree (Noise)}
    \end{subfigure}\hfill
    \begin{subfigure}[t]{0.28\linewidth}
        \centering
        \includegraphics[width=\linewidth]{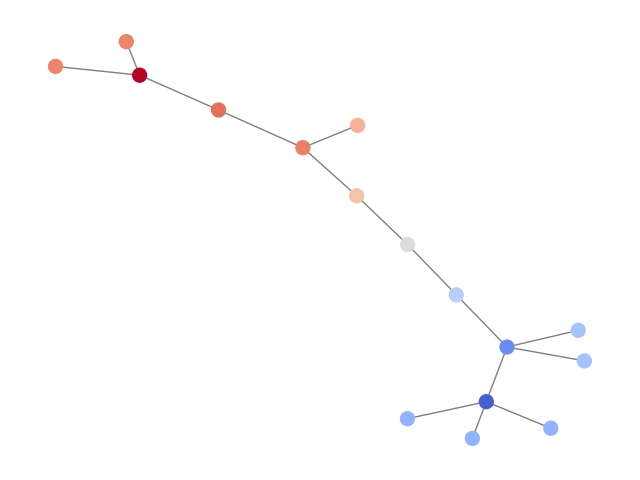}
        \caption{Tree (Coarsened)}
    \end{subfigure}\hfill
    \begin{subfigure}[t]{0.28\linewidth}
        \centering
        \includegraphics[width=\linewidth]{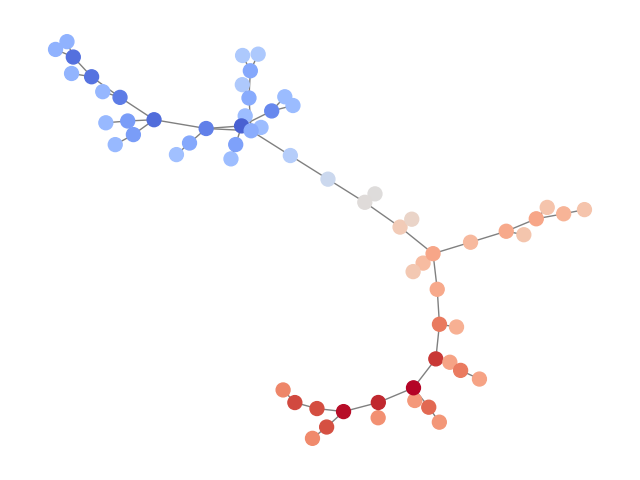}
        \caption{Tree (Expanded)}
    \end{subfigure}

    \vspace{0.4cm}
    \begin{subfigure}[t]{0.28\linewidth}
        \centering
        \includegraphics[width=\linewidth]{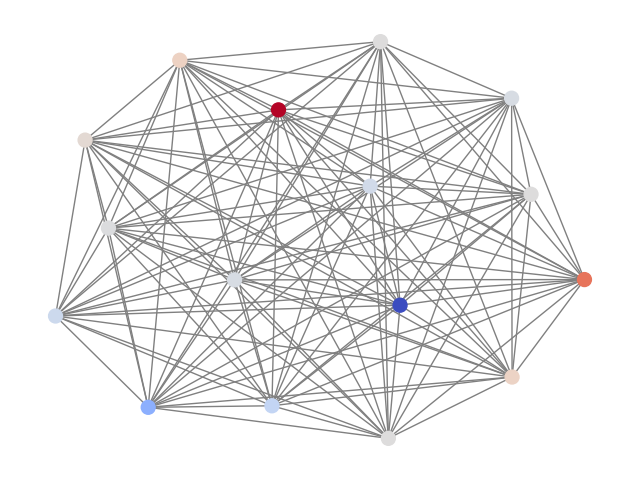}
        \caption{Planar (Noise)}
    \end{subfigure}\hfill
    \begin{subfigure}[t]{0.28\linewidth}
        \centering
        \includegraphics[width=\linewidth]{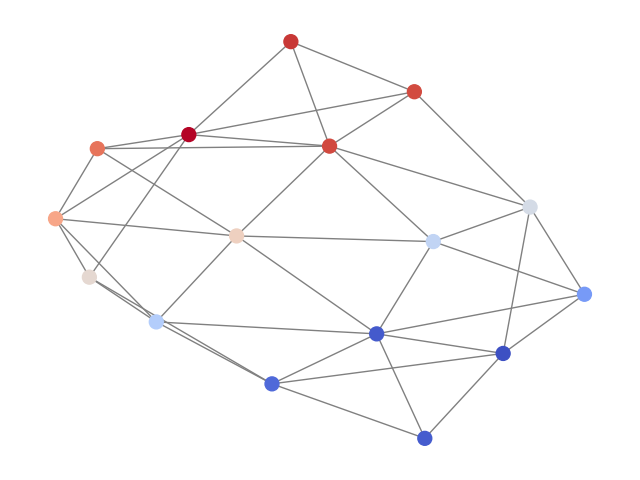}
        \caption{Planar (Coarsened)}
    \end{subfigure}\hfill
    \begin{subfigure}[t]{0.28\linewidth}
        \centering
        \includegraphics[width=\linewidth]{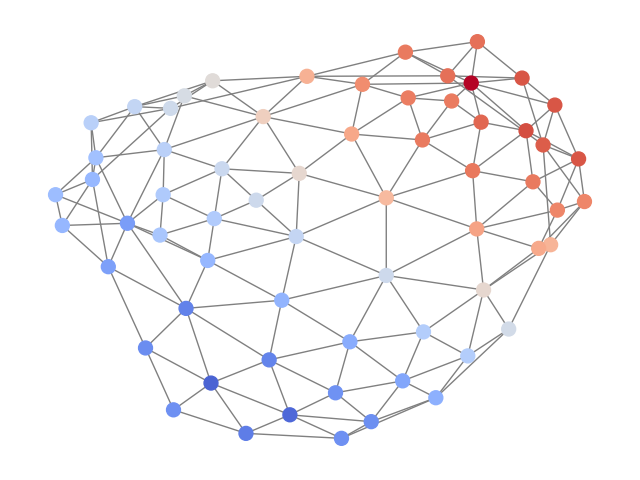}
        \caption{Planar (Expanded)}
    \end{subfigure}

    \caption{Visualization of graph generation stages across three datasets. 
    Rows correspond to datasets (Comm20, Tree, Planar), and columns correspond to stages (Noise, Coarsened, Expanded).}
    \label{fig:graph_images_grid}
\end{figure*}

\end{document}